\documentclass[11pt,a4paper]{article}
\pdfoutput=1
\usepackage{mtsummit2019}
\usepackage{color}
\usepackage{times}
\usepackage{latexsym}
\usepackage{amsmath}
\usepackage{amsfonts}
\usepackage[utf8]{inputenc}
\usepackage{amssymb}
\usepackage{multirow}
\usepackage{booktabs}
\usepackage{graphicx}
\usepackage{subcaption}
\usepackage{url}
\usepackage[ruled,linesnumbered,noend,nofillcomment]{algorithm2e}
\SetEndCharOfAlgoLine{}

  \title{Interactive-Predictive Neural Machine Translation through Reinforcement and Imitation}

\author{Tsz Kin Lam$^{\ast}$ \and Shigehiko Schamoni$^{\dagger,\ast}$ \and Stefan Riezler$^{\dagger,\ast}$ \\
  $^{\ast}$Computational Linguistics \& $^\dagger$IWR, Heidelberg University, Germany \\
  {\tt \{lam,schamoni,riezler\}@cl.uni-heidelberg.de}\\
  }

\date{}

\begin{document}
\maketitle
\begin{abstract}
  We propose an interactive-predictive neural machine translation framework 
  for easier model personalization using reinforcement and imitation learning. 
  During the interactive translation process, the user is asked for feedback 
  on uncertain locations identified by the system. 
  Responses are weak feedback in the form of ``keep'' and ``delete'' edits, 
  and expert demonstrations in the form of ``substitute'' edits. 
  Conditioning on the collected feedback, the system creates alternative 
  translations via constrained beam search. In simulation experiments on 
  two language pairs our systems get close to the performance of supervised 
  training with much less human effort. 
  \end{abstract}

\section{Introduction}
\label{sec:intro}

Despite recent success reports on neural machine translation (NMT) reaching human parity \cite{WuETAL:16,HassanETAL:18}, professional use cases of NMT require model personalization where the NMT system is adapted to user feedback provided for suggested NMT outputs \cite{WuebkerETAL:18,MichelNeubig:18}.
In this paper, we will focus on the paradigm of interactive-predictive machine translation \cite{foster1997target,BarrachinaETAL:08} which has been shown to fit easily into the sequence-to-sequence model of NMT \cite{KnowlesKoehn:16,WuebkerETAL:16}. 
The standard interactive-predictive protocol takes a human-corrected prefix as conditioning context in predicting a sentence completion, which is again corrected or accepted by the human user. Recent work showed in simulation experiments that human effort can be reduced by asking humans for reward signals or validations of partial system outputs instead of for corrections \cite{LamETAL:18,DomingoETAL:17}. 

Our goal is to combine both feedback modes --- corrections and rewards --- by treating them as expert demonstrations and reward values in an interactive protocol that combines imitation learning (IL) \cite{RossETAL:11} and reinforcement learning (RL) \cite{SuttonBarto:18}, respectively, using only limited human edits. A further difference of our framework to standard interactive-predictive NMT is our use of an uncertainty criterion that reduces the amount of feedback requests to the tokens where the entropy of the policy distribution is highest. This idea has been used successfully before in \newcite{LamETAL:18} and \newcite{PerisCasacuberta:18} and connects our work to the area of active learning \cite{SettlesCraven:08}. Lastly, our framework differs from prior work by allowing model updates based on partial translations.

Our experiments show that weak feedback in form of keep/delete rewards on translation outputs yields consistent improvements of between 2.6 and 4.3 BLEU points over the pre-trained baseline. On one language pair, it even matches the improvements gained by forcing word substitutions from reference translations into the re-decoded output. Furthermore, both feedback scenarios considerably reduce human effort.

\section{Related Work}
\label{sec:related}

Interactive-predictive translation goes back to early approaches for IBM-type \cite{foster1997target,FosterETAL:02} and phrase-based machine translation \cite{BarrachinaETAL:08,GreenETAL:14}. Knowles and Koehn \shortcite{KnowlesKoehn:16} and Wuebker et al. \shortcite{WuebkerETAL:16} presented {neural interactive translation prediction} --- a translation scenario where translators interact with an NMT system by accepting or correcting subsequent target tokens suggested by the NMT system in an auto-complete style. However, in their work the system parameters are not updated based on the prefix. This idea is implemented in \newcite{TurchiETAL:2017}, \newcite{MichelNeubig:18}, \newcite{WuebkerETAL:18}, \newcite{KarimovaETAL2018}, or \newcite{PerisETAL2017}. 
In contrast to our work, these approaches use complete post-edited sentences to update their system, while we update our model based on partial translations. Furthermore, our approach employs techniques to reduce the number of interactions. 

Our work is also closely related to approaches for interactive pre-post-editing \cite{MarieMax:15,DomingoETAL:17}. The core idea is to ask the translator to mark good segments and use these for a more informed re-decoding, while we integrate constraints derived from diverse human feedback to interactively improve decoding. Additionally, we try to reduce human effort by minimizing the number of feedback requests and by frequent model updates.

Several recent approaches to {reinforcement learning from human feedback} implement the idea of reinforcing/penalizing a targeted set of actions. 
\newcite{KreutzerETAL2018} presented an approach were ratings from human users on full translations are used successfully for NMT domain adaptation. 
Simulations of NMT systems interacting with human feedback have been presented firstly by \newcite{KreutzerETAL:17}, \newcite{NguyenETAL:17}, or  \newcite{BahdanauETAL:17}, who apply different policy gradient algorithms, William's REINFORCE \cite{Williams:92} or advantage-actor-critic methods \cite{MnihETAL:16}, respectively. In this paper, we use REINFORCE update strategies for simulated bandit feedback on the sub-sentence level.

 Gonz{\'a}lez-Rubio et al. \shortcite{Gonzalez-RubioETAL:11,Gonzalez-RubioETAL:12} apply active learning for interactive machine translation, where a user interactively finishes translations of a statistical MT system. Their active learning component decides which sentences to sample for translation and receive supervision for, and the MT system is updated on-line \cite{Ortiz-MartinezETAL:10}. In our algorithm, the active learning component decides which prefixes to receive feedback for based on the entropy of the policy distribution. 

\section{Learning Interactive-Predictive NMT from Rewards and Demonstrations}
\label{sec:learning}

As shown in \newcite{ChengETAL:18}, IL and RL can be viewed as a single algorithm that only differs in the choice of the oracle, based on objective functions that are defined as the expected value function with respect to the current model's policy $\pi_n$ in case of RL, and as the expected value function with respect to an expert policy $\pi^\ast$ in case of IL. Applied to NMT, both IL and RL are based on a Markov Decision Process where a deterministic sequence of states consisting of the source input and the history of the model's predictions (possibly incorporating expert's demonstrations) serves as conditioning context to predict the respective word, or ``action'' \cite{BahdanauETAL:17}. 

We instantiate rewards and demonstrations to the feedback types in interactive-predictive translation as follows: In the first case, uncertain words predicted by the system receive a positive or negative reward based on ``keep'' or ``delete'' feedback respectively. In the second case, uncertain words can additionally be corrected based on an expert policy in the form of ``substitute'' feedback associated with a positive reward. This feedback is integrated in context of the model's own predictions by adding rules to constrained beam search decoding \cite{HokampLiu:17,PostVilar:18}.\footnote{We observe that the distinction between weak feedback and expert feedback is difficult to make in the ``keep'' feedback case: on the one hand, this type of feedback refers to an action generated by the system, and on the other hand, it can be seen as a form of expert demonstration. From this perspective, our first system is closer to RL while our second system is closer to IL. For brevity, we will refer to our models as ``RL model'' and ``IL model'', respectively.}

\subsection{Learning Objective}

We formalize the objective of interactive-predictive NMT as maximizing the value function $V$ of a parametrized policy $\pi_{\theta}$, i.e., we seek to maximize the expected (future) reward obtainable from interactions of the NMT system with a human translator who, by editing translations, implicitly assigns rewards $R(\mathbf{\hat{y}})$ to system predictions $\mathbf{\hat{y}}$ given source sentences $\mathbf{x}$:
\begin{align}
\max_{\theta} V_{\pi_\theta}&(\mathbf{\hat{y}}; \mathbf{x}) = \max_{\theta}\mathbb{E}_{\mathbf{\hat{y}}\sim \pi_{\theta}(\cdot | \mathbf{x})}[R(\mathbf{\hat{y}})]
\label{eqn:objective}
\end{align}
Following the policy gradient theorem \cite{sutton2000policy,BahdanauETAL:17}, its derivative is
\begin{align}
\nabla_{\theta}V_{\pi_\theta}  = \mathbb{E}_{\mathbf{\hat{y}} \sim \pi_{\theta}(\cdot | \mathbf{x})} \sum_{t=1}^{T}\sum_{y\in \mathcal{V}}\nabla_{\theta}\pi_{\theta}(y | \mathbf{x},\mathbf{\hat{y}_{<t}})R(y)
\label{eqn:gradient}
\end{align}
where $\mathcal{V}$ is a vocabulary of target words. In our application, we ask for feedback on a single trajectory at each round of interactions. Similar to \newcite{Williams:92}, we consider a 1-sample estimate to reduce the inner sum of actions at each time step to the single action $\hat{y}_{t}$ presented to the user.

Depending on the type of feedback, the instantaneous reward $R(\hat{y}_{t})$ for a system translation $\hat{y}_{t}$ is set to the following values:
\begin{equation}
    R(\hat{y}_{t})= 
\begin{cases}
    0.5    & 
    \text{if } \textsc{substitute} / \textsc{keep}, \\
    -0.1  & 
    \text{if } \textsc{delete}. \\

\end{cases}
\label{eqn:reward}
\end{equation}
In addition, we found that flooring rewards for tokens that do not receive explicit feedback to a small number\footnote{We apply Gaussian noise with mean 0.1 and standard deviation of 0.05.} stabilizes the training and improves performance on the dev set. 

\section{Algorithms}
\label{sec:implementation}

\begin{figure}[t]
\centering
\includegraphics[scale=0.6]{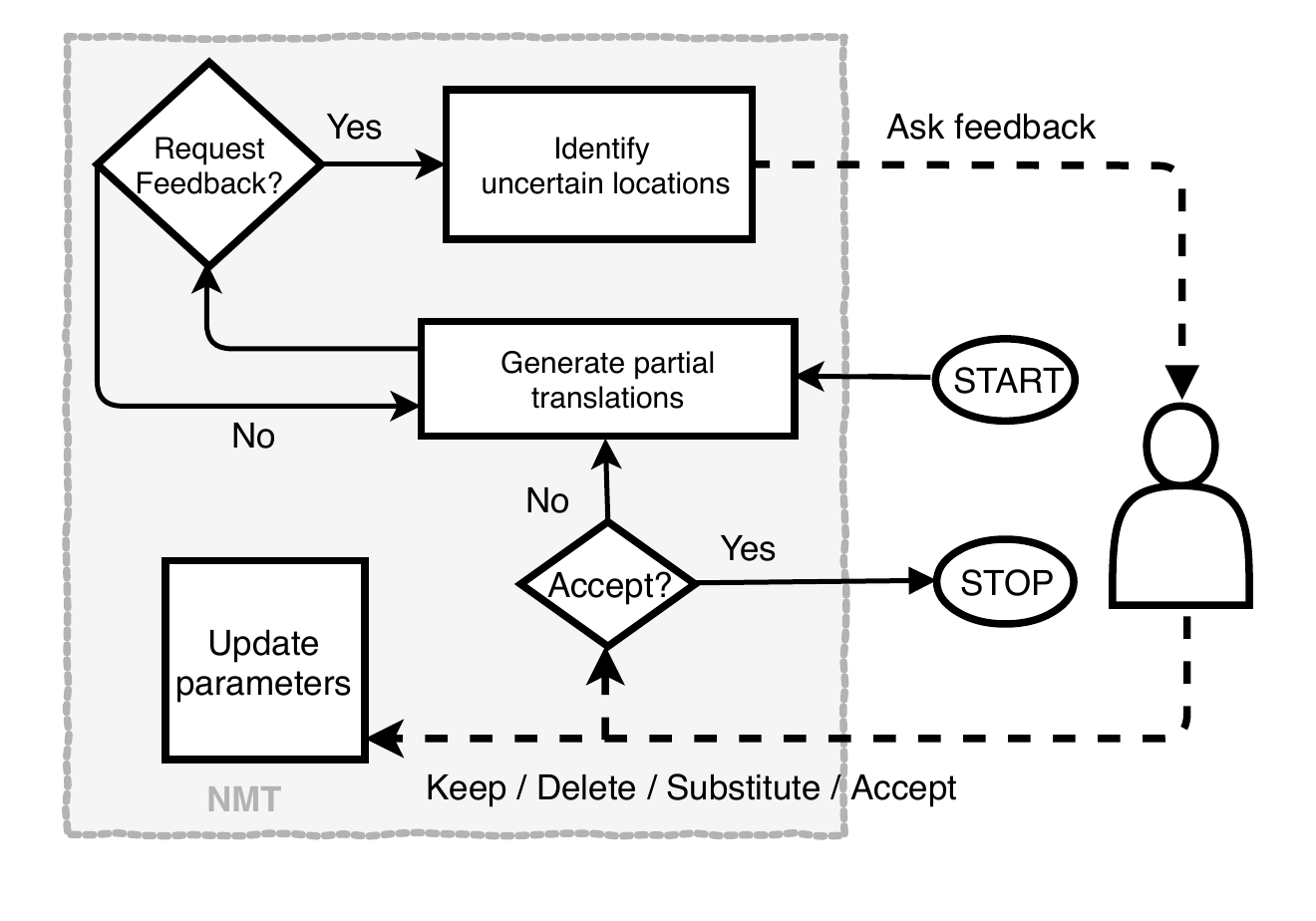}
\caption{A graphical illustration of the interactive-predictive workflow of our system. Dotted arrows indicate interactions between human and system; solid arrows indicate procedures within the system.}
\label{fig:worflow}
\end{figure}

In this section, we present the details of our interactive-predictive workflow and describe the system components of our implementation to reduce human effort while maintaining high quality model adaptation. In contrast to existing approaches where full sentences are corrected in each round, our system stops decoding when the generated segment meets several (un)certainty criteria. Our system then identifies uncertain words within the generated segment and asks the user to edit these words. The idea is to direct the user to possible translation errors in the segment, and to collect feedback on these highly informative locations, effectively implementing an active learning strategy. The collected feedback is used twice: first, it is used to perform an on-line update of the system's parameters, and secondly, it is integrated as rules into constrained beam search. 
The full translation is reached after several interactive rounds when the translator finally accepts the translation. Figure \ref{fig:worflow} gives a graphical illustration of the workflow. 

\subsection{Measuring uncertainty} 
We define a measure of uncertainty based on the entropy at a time step $t$ given a set of actions $\mathcal{V}$ (i.e., the target vocabulary) where
\begin{equation*}
H_t = -\sum_{y \in \mathcal{V}} \pi_\theta(y| \mathbf{x},\mathbf{\hat{y}_{<t}}) \log \pi_\theta(y| \mathbf{x},\mathbf{\hat{y}_{<t}}).
\end{equation*} 
The idea is that learning from edits on high entropy time steps is more helpful than learning from edits on low entropy time steps, because updating parameters based on uncertain regions better stabilizes the model over time. 
Furthermore, entropy is computationally simple and far less expensive than external reward estimators such as a quality estimation system, a critic, or a discriminator. 

A single token at time step $t$ is considered {uncertain} if the entropy exceeds a defined threshold $\epsilon$, i.e., $H_t > \epsilon$.
We use this criterion 
to identify informative locations of a partial translation on which the user is asked for feedback. 

In case of partial translations, a sequence of length $t$ is considered {uncertain} if the token at time $t$ is {uncertain} as defined above, and there is an abrupt change in entropy at $t$, formally $\frac{H_t - H_{t-1}}{H_{t-1}} > \delta$.  Both criteria are applied 
to determine the length of a partial translation shown to the user.

\begin{algorithm}[t] 
\footnotesize{
	\SetInd{0.6em}{0.6em}
	\SetKwData{K}{k}
	\SetKwFunction{IsUncertainLocation}{\textsc{uncertain-location}}
	\SetKwFunction{ConstrainedBeamSearch}{\textsc{beam-search}}
	\SetKwFunction{SetSource}{\textsc{set-nmt-source}}
	$t_{prefix} \leftarrow 1$, $n \leftarrow 1$\;
	$\theta_{0} \leftarrow \theta$, $\xi \leftarrow \emptyset$\;
	\SetSource{$\mathbf{x}$}\;
	\Repeat{$\hat{\mathbf{y}}_{1:t}$ accepted}{ 
		$\hat{\mathbf{y}}_{1:t} \leftarrow$ \ConstrainedBeamSearch{$k$, $t_{prefix}$, $T_{\max}$, $\xi$}\;
		\For{$i \leftarrow 1$ \KwTo $t$}{
			\lIf{\IsUncertainLocation{$\hat{\mathbf{y}}_{1:t},i$}}{Collect feedback rules $\xi_i$}
		}
		Get rewards for $\mathbf{\xi}_i \in \{ keep, delete, substitute \}$ according to Eq.~\ref{eqn:reward} \;
		$\theta_n \leftarrow \theta_{n-1} + \alpha \nabla_{\theta} V$ (Eq.~\ref{eqn:gradient})\;
		$t_{prefix} \leftarrow |\hat{\mathbf{y}}_{1:t}|$, $n \leftarrow n + 1$\;
	}
\caption{Interactive-predictive workflow for a single sentence using constrained beam search. \textit{Input:} model parameters $\theta$, source sentence $\mathbf{x}$,  beam size $k$, learning rate $\alpha$. \textit{Output:} updated $\theta^\ast$.}
\label{alg:intactpred}
}
\end{algorithm}

\subsection{Interactive-predictive workflow} 
Algorithm~\ref{alg:intactpred} describes the workflow in our interactive-predictive machine translation scenario. In the first round, the system starts with initial model parameters $\theta_0$, and an empty set of feedback rules $\xi$, and calls \textsc{beam-search} to first generate an unconstrained partial translation of length $t$
by evaluating the uncertainty criteria in function \textsc{is-uncertain}. The algorithm then evaluates each token within the partial translation and asks for user feedback if the token is considered uncertain 
w.r.t. the function \textsc{uncertain-location}. 

Feedback is captured in form of rules that correspond to edits on specific locations, e.g., \textsc{keep} token at position $i$, \textsc{delete} token at position $i$, or \textsc{substitute} token at position $i$ with another token.
After collecting the rewards for feedback rules $\xi_i$ according to Equation~\ref{eqn:reward}, the model parameters are updated by taking a gradient step as defined in Equation~\ref{eqn:gradient}. 

The updated system then proceeds to the next round by calling \textsc{beam-search} again, this time with a set of feedback rules $\xi$ to generate a constrained partial translation exceeding the previous length $t_{prefix}$. The uncertainty criterion of tokens is evaluated again and the user is asked for feedback on these tokens, extending the set of feedback rules $\xi$, which are used to update the system parameters and generate the next partial translation until the user is satisfied with the translation. 

\begin{algorithm}[t] 
\footnotesize{
	\SetInd{0.6em}{0.6em}
	\SetKwData{Words}{wc$_t$}
	\SetKwData{Beam}{beam}
	\SetKwData{Scores}{scores}
	\SetKwData{Distribution}{distribution$_{(y_t|\mathbf{x})}$}\SetKwData{Actions}{actions}
	\SetKwFunction{BeamSearch}{\textsc{beam-search}}
	\SetKwFunction{Kbest}{\textsc{kbest}}
	\SetKwFunction{ArgmaxK}{\textsc{argmax}$_k$}
	\SetKwFunction{DecoderInit}{\textsc{decoder-init}}
	\SetKwFunction{DecoderStep}{\textsc{decoder-step}}
	\SetKwFunction{IsUncertain}{\textsc{is-uncertain}}
	\SetKwFunction{ApplyConstraints}{\textsc{apply-constraints}}
	\SetKwFunction{Length}{\textsc{length}}
	\SetKwProg{Fn}{function}{}{}
\Fn{\BeamSearch{$k$, $p$, $N$, $\xi$}}{
	$beam$ $\leftarrow$ \DecoderInit{k}\;
	\For{$t \leftarrow 1$ \KwTo $N$}{
		$scores \leftarrow$ \DecoderStep{$beam$}\;
		$beam \leftarrow$  \Kbest{$scores$, $k$, $\xi$}\;
		\lIf{\Length{$beam[0]$}$>p$ {\bf and} \IsUncertain{$beam[0]$}}{ break } %
	}
	\Return$beam[0]$\;
}

\Fn{\Kbest{$scores$, $k$, $\xi$}}{
	$scores_c$ $\leftarrow$ \ApplyConstraints{$scores$, $\xi$}\;
	$beam$ $\leftarrow$ \ArgmaxK{$scores_c$}\;
	\Return $beam$\;

}
	\caption{Constrained beam search for uncertain partial translation. \textit{Input:} beam size $k$, prefix length $p$, maximum length $N$, feedback rules $\xi$. \textit{Output:} partial translation.}
\label{alg:constrainedbeamsearch}
}
\end{algorithm}

\subsection{Constrained beam search} 
\label{sec:constrainedbeamsearch}
A central component is a modified beam search algorithm that takes positional constraints into account (Algorithm~\ref{alg:constrainedbeamsearch}). 
The user constraints force the system to generate alternative translations and can thus be interpreted as an exploration strategy. 
An efficient alternative exploration strategy is multinomial sampling. In our interactive-predictive scenario, however, it is crucial that translations on locations without explicit user feedback are preserved, and this cannot be modeled easily with multinomial sampling. Beam search on the other hand ensures stable translations due to its deterministic nature, and the idea of constrained beam search provides the tools to improve the translation interactively. As a side effect, higher quality translations can be obtained by increasing the beam size at the cost of computational power. 

After initializing $k$ beams, 
the algorithms generates a partial translation by calling \textsc{decoder-step} to retrieve the next token and score all hypotheses. The constraints (provided in the form of feedback rules) are applied in the function \textsc{kbest} by filtering out all hypotheses that do not satisfy the constraints before the $\textsc{argmax}_k$ operation selects the $k$ highest scoring remaining hypotheses. 
The single best partial translation is shown to the user only if two conditions are met: (1) the length exceeds the length of the previous partial translation, and (2) the current partial translation is considered an {uncertain sequence}. In case one condition is not met, the system iteratively extends the partial translation up to a maximum hypothesis length.

\section{Experiments}
\label{sec:experiments}

To demonstrate the effectiveness of our reinforcement and imitation strategies, we simulate the interactive-predictive workflow described in Section~\ref{sec:implementation} in a domain adaptation setup. A human translator is simulated by comparing partial translations with corresponding gold translation to extend the set of feedback rules in every round. In the RL setting, the simulated human translator provides only weak feedback (\textsc{keep} and \textsc{delete} edits) on tokens generated by the system, while in the IL setting the simulated translator additionally injects expert feedback (\textsc{substitute} edit) by demonstrating how the system should act at a specific time step. 
In our simulation experiments, we focus on the uncertain tokens of the partial translation. 
An exact match between the uncertain token and the reference generates a \textsc{keep} edit, while differing tokens generate either a \textsc{delete} or \textsc{substitute} edit depending on the type of system. Tokens exceeding the sentence length of the reference always receive a \textsc{delete} feedback. We refer to the first system as \textsc{keep+delete}, and the second system as \textsc{+substitute}.
While the system parameters are updated online after every such simulated interaction, system evaluation is done by a standard offline translation of an unseen test set.

\begin{table}[t]
\centering
\resizebox{\columnwidth}{!}{

\begin{tabular}{lcccc}
\toprule
& Data & Training & train / dev / test & $\varnothing$ en-length\\
 \midrule
\multirow{2}{*}{\rotatebox[origin=c]{90}{fr-en}} & EP & pre-training & 1.3M / 2k / -- & 25.5\\
&NC  & interactive & 18.4k / 3k / 5k  & 22.8 \\
 \midrule
\multirow{2}{*}{\rotatebox[origin=c]{90}{de-en}} & EP & pre-training &1.7M / 2.7k / -- & 24.0\\
&NC  & interactive & 18.9k / 1k / 2k  & 22.6\\


\end{tabular}
}
\caption{Data used in pre- and interactive training for French-English (fr-en) and German-English (de-en).}
\label{tab:datasets}
\end{table}

\subsection{Dataset} 

For pre-training, we use the Europarl (EP) corpus version 5 for the French-English system, and version 7 for German-English. 
For interactive training, we use the News Commentary (NC) 2006 corpus. 
Both corpora are publicly available on the WMT13's homepage.\footnote{\url{https://www.statmt.org/wmt13/}} 
All experiments are conducted on two language pairs, i.e., German-English (de-en) and French-English (fr-en). Data sets were tokenized and lowercased using \textsc{Moses} preprocessing scripts \cite{koehn2007moses}. We applied compound splitting on the German source sentences using \textsc{cdec}'s tool \cite{dyer2010cdec}. 
Our data sets for interactive training differ from the original News Commentary data splits as follows:
(1) we sample a subset of the original training set to reduce the number of parallel sentences to 18,432 for French-English and 18,927 for German-English, and (2) we increase both validation and test set for French-English to 3,001 and 5,014 parallel sentences by moving data from the original training set excluding sentences that were sampled for training. 
Note that a training set size of less than 19,000 parallel sentences is very small even in a domain adaptation setup. Table~\ref{tab:datasets} summarizes the statistics of our datasets. 

\begin{table*}[t!]
\setlength{\tabcolsep}{4pt} 
\centering

\begin{tabular}{lrcccccc}
\toprule
Pair & System & ChrF ($\sigma$) & $\Delta$ChrF & BLEU ($\sigma$) & $\Delta$BLEU & $\varnothing$ rounds & $\varnothing$ keep+delete / subst. \\
 \midrule
\multirow{4}{*}{\rotatebox[origin=c]{90}{\textbf{fr-en}}} & Pre-trained & 61.08 & -- & 24.70 & -- &  -- & -- \\
& Full Post Edits & 61.96 (0.15) & +0.88  & 29.10 (0.09) & +4.40 & -- & --\\
& \textsc{keep}+\textsc{delete} & \textbf{62.72} (0.11) & +1.64  & 28.16 (0.14) & +3.46 & 3.2 & 13.7 / -- \\
& +\textsc{substitute} & 62.24 (0.08) & +1.16 & \textbf{28.52} (0.10) & +3.82 & 3.3 & 1.8 / 5.6 \\
 \midrule
\multirow{4}{*}{\rotatebox[origin=c]{90}{\textbf{de-en}}} & Pre-trained & 59.34 & -- & 22.66 & -- & -- & -- \\
& Full Post Edits & 60.24 (0.25) & +0.9 & 27.40 (0.22) & +4.74 & -- & --\\
&\textsc{keep}+\textsc{delete} & 59.57 (0.19) &+0.23 & 25.28 (0.09) &+2.62 & 3.3 & 13.1 / -- \\
&+\textsc{substitute} & \textbf{60.73} (0.14) &+1.39 & \textbf{26.91} (0.1) &+4.25 & 3.3 & 1.8 / 5.9 \\
\bottomrule
\end{tabular}
\caption{Character-F (ChrF), and BLEU test results on the French-English (fr-en) and German-English (de-en) translation tasks. Highest scores on RL and IL systems are printed in bold. The $\Delta$ columns indicate the score differences to the pre-trained baseline system. All scores are averaged over three runs with standard deviation $\sigma$ in parentheses.}
\label{tab:results}
\end{table*}

\subsection{Model Architecture}

We use a single uni-directional LSTM layer with global attention mechanism between encoder and decoder. The dimensionality of the LSTM hidden states and the word embeddings are 500. We build the vocabulary using the most frequent 50,000 words in each language. 

The Adam optimizer \cite{kingma2014adam} is used in all training scenarios. In supervised training, we use a mini-batch size of 64 and an initial learning rate of 0.001. Starting from the 5$^{th}$ epoch, the rate is reduced by half in each epoch if the validation perplexity increases.
In interactive training, we train for a single epoch and apply a constant learning rate of 10$^{-5}$ with a mini-batch size of 1.
In all experiments we set entropy parameters to $\epsilon =$ 1, $\delta = $ 0.5, and use a beam size of 5 during training. For testing, we apply greedy decoding. PyTorch code of our models is publicly available.\footnote{\url{https://github.com/heidelkin/IPNMT_RL_IL}}

\subsection{Results and Discussion}

On both language pairs, the optimal pre-trained NMT models are obtained in the 6$^{\text{th}}$ training epoch, forming the out-of-domain baseline. We also compare our RL/IL strategies with full post-edits simulated by supervised training on the in-domain News Commentary data, forming an in-domain upper bound. We repeated each experiment three times and report mean and standard deviation for both Character-F\footnote{Using parameters $ngram$ = 6 and $\beta$ = 2.}  (ChrF) \cite{popovic2015chrf} 
and corpus BLEU \cite{papineni2002bleu}.

In the French-English experiments, both our imitation and reinforcement strategies show improvements of more than 3 points in BLEU and 1 point in ChrF over the out-of-domain baseline. Both strategies achieve lower BLEU score than training on full post-edits, in particular, 0.94 points lower in the \textsc{keep}+\textsc{delete} setting, and 0.58 points lower in +\textsc{substitute} setting. However, both strategies achieve higher ChrF scores, i.e., 0.76 points for \textsc{keep}+\textsc{delete} and 0.28 points for +\textsc{substitute}. See upper half of Table~\ref{tab:results} for a summary.

In the German-English experiments, there is a bigger performance gap  between the \textsc{keep}+\textsc{delete} and the full post-edits system, concretely, 0.64 points in ChrF score and 2.12 points in BLEU lower than full post-edits. However, the improvement over the pre-trained model amounts to 2.62 BLEU points and 0.25 points in ChrF score. Our +\textsc{substitute} system is comparable in performance to the full post-edits system, yielding a result that is 0.49 lower in BLEU but 0.49 points higher in ChrF. See lower half of Table~\ref{tab:results} for the summary.

We also report average numbers of feedback rounds and rules per sentence in Table~\ref{tab:results}. 
We optimized the maximum number of allowed feedback rules per round on the dev set and use 9 (fr-en) and 7 (de-en) for the \textsc{keep}+\textsc{delete} and 3 for the +\textsc{substitute} systems. Even for the simpler model based on only weak feedback, the number of user clicks is between 13.7 and 13.1, which is well below the average target sentence length of 22.8 and 22.6. By allowing expert \textsc{substitute} feedback that actively generates better tokens in the next round 
the number of rules is reduced to 7.4 and 7.7. Our experiments indicate that focusing on uncertain locations can reduce human translation effort substantially. 

\begin{figure}[t]
	\includegraphics[angle=-90,width=\columnwidth]{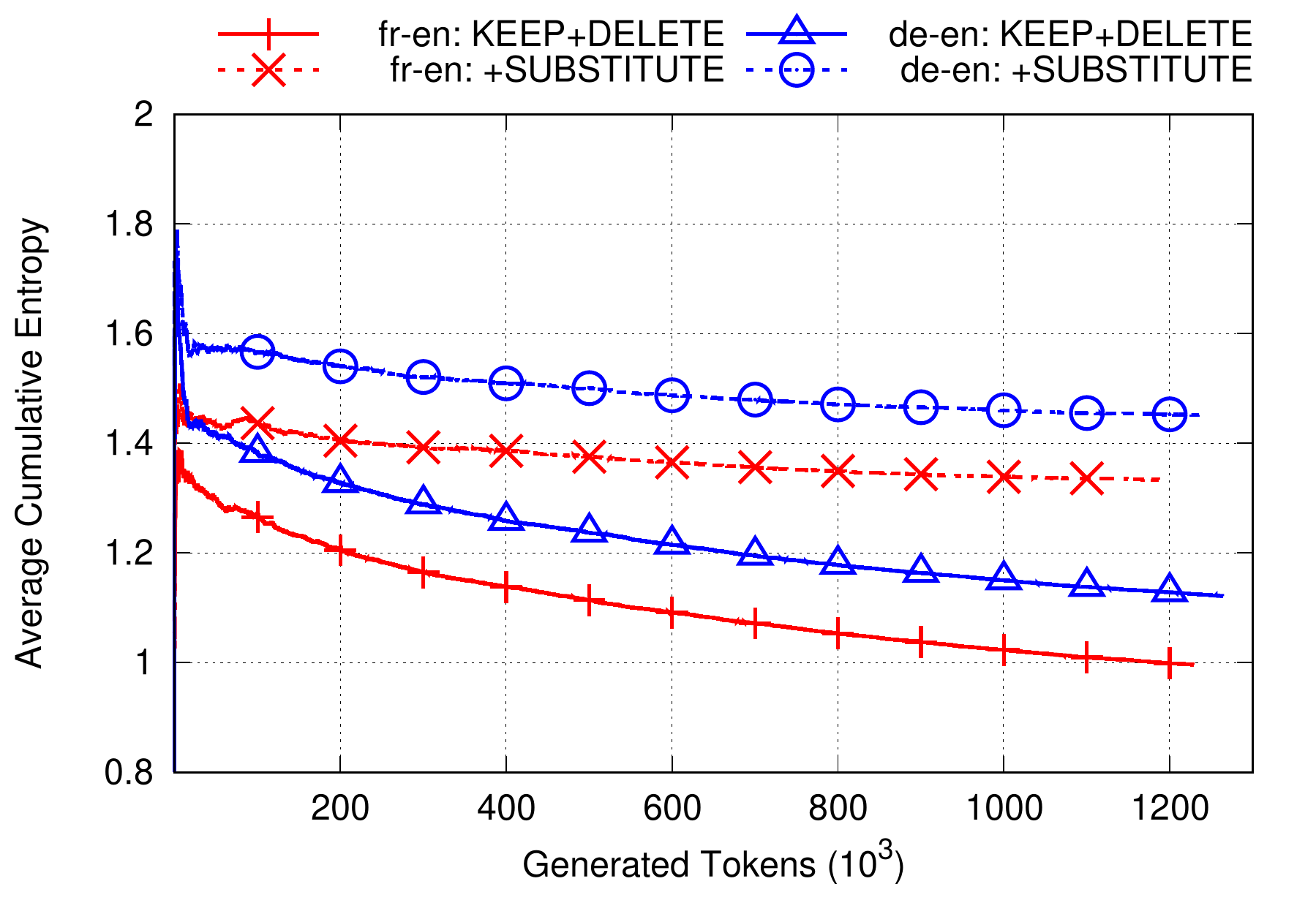}
	\caption{Average cumulative entropy of the model's policy distribution over time during simulated interactive learning. Plots are shown for the French-English ({\color{red}fr-en}) and the German-English ({\color{blue}de-en}) task, and for the \textsc{keep+delete} and the \textsc{+substitute} system, respectively.
	}
	\label{fig:entropy}
\end{figure}

\paragraph{Effect of on-line learning.} 
We also examine the effect of on-line learning on average cumulative entropy of the model's policy distribution over time. Figure \ref{fig:entropy} visualizes the change of entropy during interactive training. At the beginning, the system is in regions of high entropy but quickly learns from human edits and the curves become smooth and monotonic. 
After this initial phase, the overall better performing French-English task shows consistently lower entropy than the German-English task, indicating a connection between model's entropy and translation quality. 
However, the comparison between the \textsc{keep}+\textsc{delete} and the better performing +\textsc{substitute} systems shows the opposite trend and requires a different explanation. We conjecture that the +\textsc{substitute} system's expert demonstrations at uncertain locations help the system to find better translations, but such demonstrations also move the system to higher entropy regions, effectively implementing a useful exploration strategy. In contrast to this, the \textsc{keep}+\textsc{delete} system always stays in more certain regions by selecting another high probability token if the original token receives a \textsc{delete} feedback by the user.

\begin{figure}[t]
  \centering
    \begin{subfigure}[b]{\columnwidth}
		\includegraphics[angle=-90,width=\columnwidth]{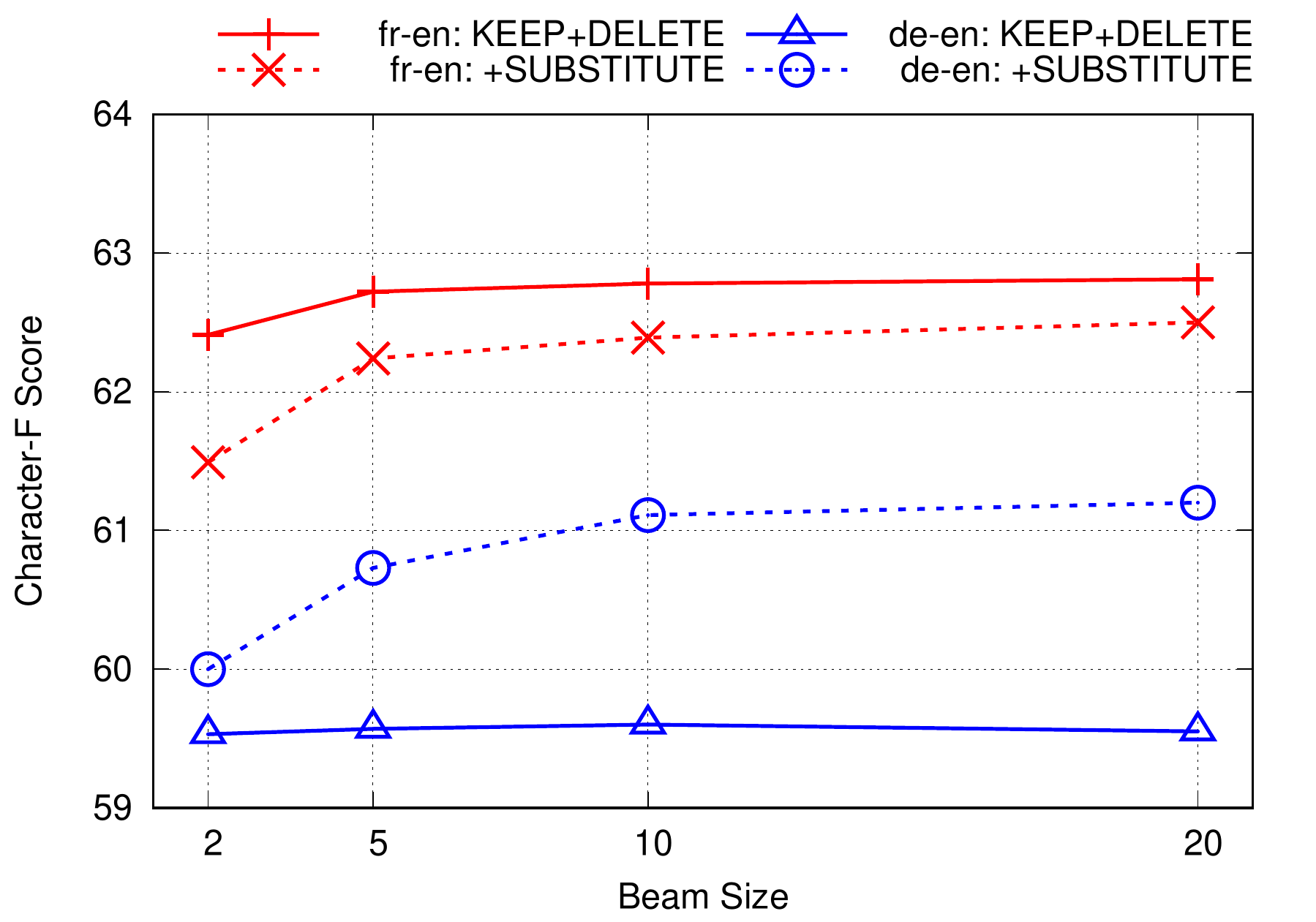}
	\end{subfigure}\\
	
    \begin{subfigure}[b]{\columnwidth}
		\includegraphics[angle=-90,width=\columnwidth]{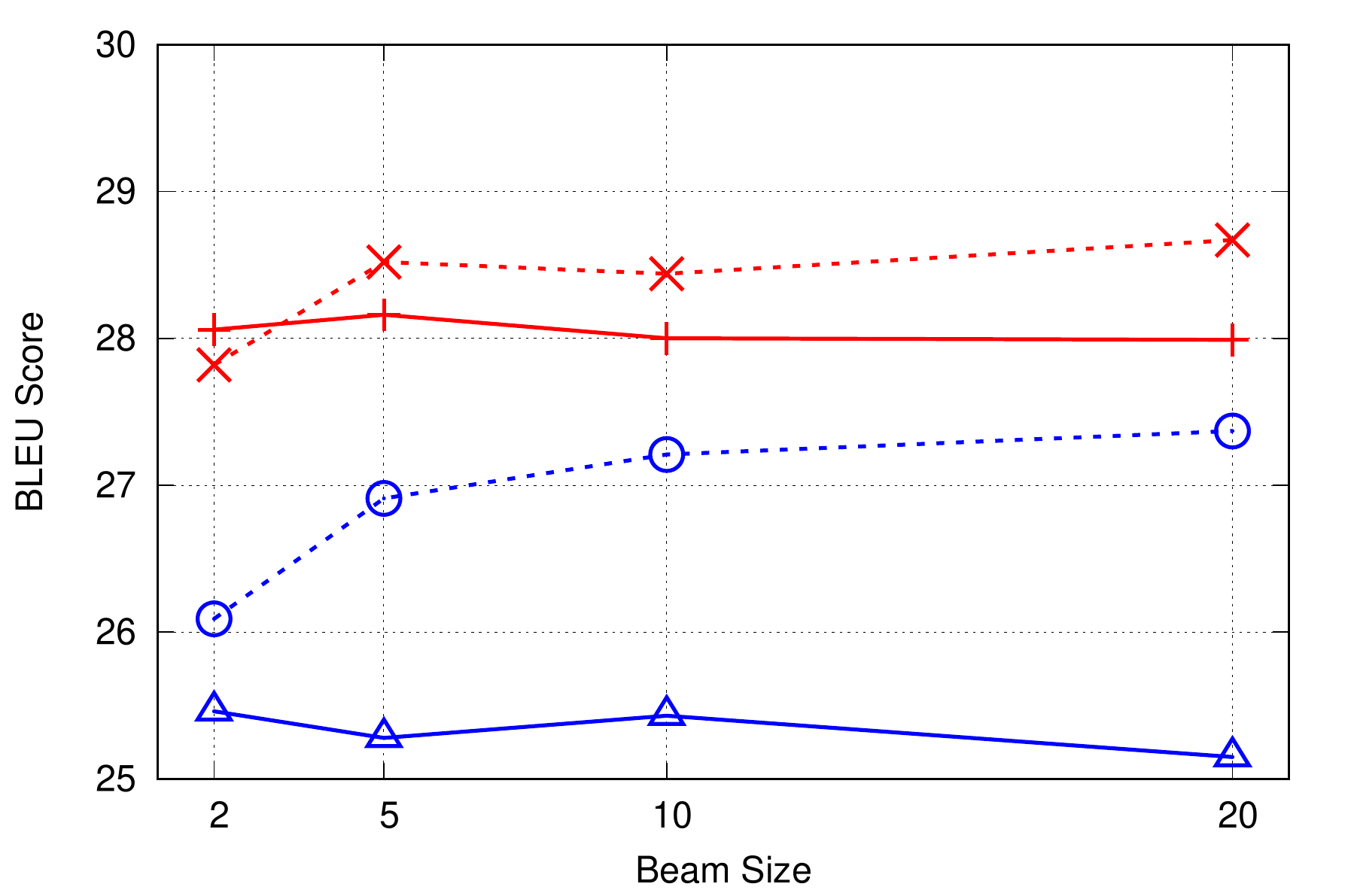}
	\end{subfigure}
\caption{The two figures show the effect of different beam sizes on Character-F score (top) and BLEU score (bottom). 
We conduct experiments on French-English ({\color{red}fr-en}) and German-English ({\color{blue}de-en}) and both systems (\textsc{keep+delete} and \textsc{+substitute}). All scores are averaged over two runs.}
\label{fig:beam_size}
\end{figure}

\paragraph{Effect of beam size.} 
The observations on model's entropy over time in the previous paragraph and the implementation details described in Section~\ref{sec:constrainedbeamsearch} show that our constrained beam search implements exploration in a user-controlled manner. We conjecture that beam size also influences the exploration and should have a different effect on different feedback strategies. We thus conduct additional experiments using beam sizes of 2, 5, 10 and 20 on all language pairs and the two systems. The results are summarized in Figure \ref{fig:beam_size}. 
In both \textsc{keep}+\textsc{delete} and +\textsc{substitute} systems, a beam size of 2 is sufficient to achieve substantial gains over the baselines in both language pairs. 
In case of the \textsc{keep}+\textsc{delete} system, increasing beam sizes only marginally influence the translation performance.
In case of the +\textsc{substitute} system, there are considerable gains of almost 1 BLEU point and 1 Character-F point when increasing the beam size from 2 to 5. 
Here, the larger beam size enables the system to connect the expert demonstrations with better prefixes which helps the system to explore higher scoring trajectories. 
Increasing the beam size to 10 or 20 further improves performance but the gains are small.

%


\begin{table*}[h!]
\centering
\resizebox{\textwidth}{!}{%

\begin{tabular}{ccl}
\toprule
\multirow{14}{*}{\rotatebox[origin=c]{90}{{German-English}}} & \textbf{Source} & {{der kern des problems ist nicht die gesamt$_\smile$menge des öls , sondern seine lage .}}\\
&\textbf{Reference} & {{the heart of the problem is not the overall quantity of oil , but its location .}} \\
\\
&\textbf{Round} & \textbf{Partial translation}  $\rightarrow$ \textbf{\textsc{feedback}} \\
& 1 & the {\color{blue}core$_2$} \\ 
  &  &  $\rightarrow$ \textsc{delete}(2)\\
  &2 & the {\color{blue}\textit{heart}$_2$} of the problem is not the {\color{blue}total$_9$}\\
  &  &  $\rightarrow$ \textsc{keep}(2),  $\rightarrow$ \textsc{delete}(9)\\
  &3 & the \textit{heart} of the problem is not the \textit{overall} {\color{blue}amount$_{10}$} of oil , but {\color{blue}its$_{15}$} \\
  &  &  $\rightarrow$ \textsc{delete}(10), $\rightarrow$ \textsc{keep}(15)\\
  &4 & the \textit{heart} of the problem is not the \textit{overall} {\color{blue}volume$_{10}$} of oil , but \textit{its} {\color{blue}situation$_{16}$} . \\  
  &  &  $\rightarrow$ \textsc{delete}(10, 16)\\
  &5 & the \textit{heart} of the problem is not the \textit{overall} \textit{supply} of oil , but \textit{its} \textit{position} . \\  
  &  &  $\rightarrow$ accepted.\\
\midrule
\multirow{12}{*}{\rotatebox[origin=c]{90}{{German-English}}} & \textbf{Source} &{{die süd$_\smile$koreaner ihrerseits verlassen sich darauf , dass china mit der nuklearen krise in nord$_\smile$korea fertig wird .}}\\
 & \textbf{Reference} & {{as for the south koreans , they are counting on china to deal with the north korean nuclear crisis . }} \\
\\
 & \textbf{Round} & \textbf{Partial translation}  $\rightarrow$ \textbf{\textsc{feedback}} \\
 & 1 & {\color{blue}the$_1$} south koreans {\color{blue}are$_4$} \\
 &   & $\rightarrow$ \textsc{substitute}(1:as, 4:south)\\
 & 2 & \textit{as} for {\color{blue}the$_3$} \textit{south} koreans , {\color{blue}china$_7$} \\
 &   & $\rightarrow$ \textsc{keep}(3:the), $\rightarrow$ \textsc{substitute}(7:they)\\
 & 3 & \textit{as} for \textit{the} \textit{south} koreans , \textit{they} are {\color{blue}relying$_9$} on china to {\color{blue}be$_{13}$} \\
 &   & $\rightarrow$ \textsc{substitute}(9:counting, 13:deal) \\
 & 4 & \textit{as} for \textit{the} \textit{south} koreans , \textit{they} are \textit{counting} on china to \textit{deal} with the nuclear crisis in north korea .\\
 &   &  $\rightarrow$ accepted.\\
\midrule
\multirow{14}{*}{\rotatebox[origin=c]{90}{{French-English}}} & \textbf{Source} & {il est dur d' aimer ou de respecter un peuple et de haïr son état .}\\
 & \textbf{Reference} & {it is hard to love or respect a people and hate their state . }\\
\\
 & \textbf{Round} & \textbf{Partial translation} $\rightarrow$ \textbf{\textsc{feedback}} \\
 & 1 & it is hard to {\color{blue}love$_5$} \\
 &   & $\rightarrow$ \textsc{keep}(5)\\
 & 2 & it is hard to \textit{love} or {\color{blue}to$_7$} \\
 &   &  $\rightarrow$ \textsc{delete}(7)\\
 & 3 & it is hard to \textit{love} or {\color{blue}\textit{comply}$_7$} with a people and {\color{blue}to$_{12}$} {\color{blue}hate$_{13}$} {\color{blue}their$_{14}$} \\
 &   &  $\rightarrow$ \textsc{delete}(7, 12, 13, 14) \\
 & 4 & it is hard to \textit{love} or {\color{blue}\textit{respect}$_7$} {\color{blue}a$_8$} people and {\color{blue}hatred$_{11}$} {\color{blue}.$_{12}$} \\
 &   &  $\rightarrow$ \textsc{keep}(7, 8),$\rightarrow$ \textsc{delete}(11, 12).\\
 & 5 & it is hard to \textit{love} or \textit{respect} \textit{a} people and \textit{to} \textit{hate} \textit{their} \textit{state} . \\
 &   & $\rightarrow$ accepted. \\
\midrule
\multirow{12}{*}{\rotatebox[origin=c]{90}{{French-English}}} & \textbf{Source} & {un gouvernement qui n' est pas en mesure d' \'{e}quilibrer ses propres finances ne peut pas apporter une stabilité macroéconomique .}\\
 & \textbf{Reference} & {a government that cannot balance its own finances cannot be relied on to provide macroeconomic stability . }\\
\\
 & \textbf{Round} & \textbf{Partial translation} $\rightarrow$ \textbf{\textsc{feedback}} \\
 & 1 & a government that {\color{blue}is$_4$} \\
 &   & $\rightarrow$ \textsc{substitute}(4:cannot)\\
 & 2 & a government that \textit{cannot} balance its {\color{blue}own$_7$} \\
 &   &  $\rightarrow$ \textsc{keep}(7)\\
 & 3 & a government that \textit{cannot} balance its \textit{own} finances cannot {\color{blue}bring$_{10}$ about$_{11}$ macro-economic$_{12}$} stability .\\
 &   &  $\rightarrow$ \textsc{substitute}(10:be,11:relied,12:on) \\
 & 4 & a government that \textit{cannot} balance its \textit{own} finances cannot \textit{be} \textit{relied} \textit{on} to bring about macro-economic stability .\\
 &   &  $\rightarrow$ accepted.\\
\bottomrule
\end{tabular}%
}
\caption{Interaction protocol illustrating translation progress of the two learning systems on the German English task (upper half) and French-English (lower half). For each language pair, the first example illustrates interactions with the 
\textsc{keep+delete} system, while the second example shows interactions with the \textsc{+substitute} system. 
In each round, the user is asked for feedback on uncertain locations of the current partial translation. 
Tokens printed in {\color{blue}blue} with their position in subscript indicate uncertain locations. At the end of each round, the system is updated given the user's feedback (\textsc{keep}, \textsc{delete}, \textsc{substitute}). In the next round, it generates a constrained (partial) translation with respect to this feedback. Tokens generated based on feedback rules are printed in \textit{italics}.
 }
\label{tab:examples}
\end{table*}


\paragraph{Decoding Speed.}

The total runtime of each of our simulated interactive experiments is roughly 6 hours when simulated on a Nvidia P40, while training of the  \textsc{keep}+\textsc{delete} system is slightly slower than of the +\textsc{substitute} system due to the higher number of feedback rules. Looking at the sentence level this means the total decoding time of our system for all partial translations of a single sentence is $6 \times 1\textrm{h} / (18,432 \times 3.3) = 0.361\textrm{s}$ for the French-English task, and even less for the German-English task. This estimate does not account for the time our system conducts validation tests or constructs simulated feedback, thus the actual average processing time is lower. 
\newcite{KnowlesKoehn:16} argue that beam search is usually too slow to be used for training in interactive live systems, however, recent hardware developments together with our strategy of partial decoding makes constrained beam search applicable even in training. As a side effect, corrections on early time steps reduce the problem of error propagation and thus improve both usability of the system and satisfaction of the translator. 

\paragraph{Leveraging BPE or character-level NMT.} Our current implementation of interactive-predictive NMT uses a word-based translation approach and presents word units to users for feedback. An adaptation of our algorithm to sub-word or  character level NMT is possible and requires to redistribute the reward associated to the word level to sub-word units or characters, and to maintain their location information in the constrained beam search. We leave this extension to future work.

\subsection{Examples}

Table \ref{tab:examples} illustrates the translation workflow of our interactive-predictive protocol by listing four examples: the upper half shows example translations of the two systems for the German-English task, the lower half shows two examples of the systems for the French-English task. 

The first example is taken from the \textsc{keep+delete} system, where our simulated user provides only \textsc{keep} and \textsc{delete} feedback on suggested locations. In interactive round 1 on the German-English task, the system stops after generating the uncertain partial translation ``the core'' and asks the user for feedback specifically on the term ``core''. The simulated user returns a \textsc{delete} feedback and the system is able to generate the more appropriate translation ``heart of the problem'' in round 2. In round 3, however, a weakness of the simulated feedback becomes apparent: our user gives a negative \textsc{delete} feedback on the token ``amount'' because the token differs from the given reference word ``quantity'', even though it is an appropriate translation for the German word ``Menge'' in this context. The system then generates ``volume'' in round 4 and ``supply'' in the final round 5, although both translations are worse than the initially proposed translation ``amount''. One explanation for this behavior is the way on-line updates are applied to the NMT system: while the constrained beam search implements feedback rules on token level, the on-line updates of the NMT system take place on the word embedding level. 
An update based on negative feedback actually forces the NMT system to avoid semantically similar words. In the above example, the negative feedback for ``amount'' downgrades the optimal translation ``quantity'' because of the semantic similarity of both words, and instead upgrades the more diverse translations ``volume'' and ``supply''. In our example, this strategy has an immediate negative impact on translation quality, but it also illustrates the positive exploration effect which is helpful in the long run. 

The second example is taken from the \textsc{+substitute} system, where the simulated user additionally provides ``substitute'' feedback. 
In interactive round 1, the system generates the uncertain partial translation ``the south koreans are'' and identifies ``the'' and ``are'' as uncertain tokens. The user suggests to change ``the'' to ``as'', and ``are'' to ``south'' by providing \textsc{substitute} feedback. Again, a limitation of our simulation becomes apparent: our simulated substitutions are based on reference translations, but a real translator would not change the given partial translation to ``as south korean south''. Still, based on the two feedback rules and the on-line update, the NMT system is able to follow a better trajectory in round 2. We observe that \textsc{substitute} feedback is a very strong signal that supports the system to quickly get close to the translation our simulated user has in mind (which is the reference in our simulation). 

The French-English task examples illustrate a noteworthy property of our algorithm: In round 3 of the \textsc{keep+delete} system, the simulated user provides \textsc{delete} feedback on the tokens ``to hate their'' only because they occur at different positions compared to the reference. However, the system is able to recover and re-generate the tokens at the correct position in round 5. A similar behavior can be observed for the \textsc{+substitute} system in round 3, where the phrase ``bring about macro-economic'' is first substituted and then generated again in the final round 4.

\section{Conclusion}
\label{sec:conclusion}

In this work, we integrate interactive-predictive NMT with imitation learning and reinforcement learning. 
Our goal is to merge the human edit process with effort reduction and model learning into a single framework for easier model personalization. 
Our results indicate that on-line learning from edits on uncertain locations of partial translations can achieve performance comparable to using supervised learning on in-domain data but with substantially less human effort. In the future, we would like to investigate the limitations of entropy-based uncertainty measures, work on the efficiency of the training speed, 
and conduct field studies with human users.

\section*{Acknowledgments.} We would like to thank the anonymous reviewers for their feedback. The research reported in this paper was supported in part by the German research foundation (DFG) under grant RI-2221/4-1.

\bibliography{paper_arxiv}

\begin{thebibliography}{}

\bibitem[\protect\citename{Bahdanau \bgroup et al.\egroup
  }2017]{BahdanauETAL:17}
Bahdanau, Dzmitry, Philemon Brakel, Kelvin Xu, Anirudh Goyal, Ryan Lowe, Joelle
  Pineau, Aaron Courville, and Yoshua Bengio.
\newblock 2017.
\newblock An actor-critic algorithm for sequence prediction.
\newblock In {\em Proceedings of the 5th International Conference on Learning
  Representations {(ICLR)}}, Toulon, France.

\bibitem[\protect\citename{Barrachina \bgroup et al.\egroup
  }2008]{BarrachinaETAL:08}
Barrachina, Sergio, Oliver Bender, Francisco Casacuberta, Jorge Civera, Elsa
  Cubel, Shahram Khadivi, Antonio Lagarda, Hermann Ney, Jes{\'u}s Tom{\'a}s,
  Enrique Vidal, and Juan-Miguel Vilar.
\newblock 2008.
\newblock Statistical approaches to computer-assisted translation.
\newblock {\em Computational Linguistics}, 35(1):3--28.

\bibitem[\protect\citename{Cheng \bgroup et al.\egroup }2018]{ChengETAL:18}
Cheng, Ching-An, Xinyan Yan, Nolan Wagener, and Byron Boots.
\newblock 2018.
\newblock Fast policy learning through imitation and reinforcement.
\newblock In {\em Uncertainty in Artificial Intelligence {(UAI)}}, Monterey,
  {CA, USA}.

\bibitem[\protect\citename{Domingo \bgroup et al.\egroup }2017]{DomingoETAL:17}
Domingo, Miguel, {\'A}lvaro Peris, and Francisco Casacuberta.
\newblock 2017.
\newblock Segment-based interactive-predictive machine translation.
\newblock {\em Machine Translation}, 31(4):163--185.

\bibitem[\protect\citename{Dyer \bgroup et al.\egroup }2010]{dyer2010cdec}
Dyer, Chris, Jonathan Weese, Hendra Setiawan, Adam Lopez, Ferhan Ture, Vladimir
  Eidelman, Juri Ganitkevitch, Phil Blunsom, and Philip Resnik.
\newblock 2010.
\newblock cdec: A decoder, alignment, and learning framework for finite-state
  and context-free translation models.
\newblock In {\em Proceedings of the ACL 2010 System Demonstrations (ACL
  Demo)}, Uppsala, Sweden.

\bibitem[\protect\citename{Foster \bgroup et al.\egroup
  }1997]{foster1997target}
Foster, George, Pierre Isabelle, and Pierre Plamondon.
\newblock 1997.
\newblock Target-text mediated interactive machine translation.
\newblock {\em Machine Translation}, 12(1-2):175--194.

\bibitem[\protect\citename{Foster \bgroup et al.\egroup }2002]{FosterETAL:02}
Foster, George, Philippe Langlais, and Guy Lapalme.
\newblock 2002.
\newblock User-friendly text prediction for translators.
\newblock In {\em Proceedings of the Conference on Empirical Methods in Natural
  Language Processing {(EMNLP)}}, Philadelphia, {PA}.

\bibitem[\protect\citename{Gonz{\'a}lez-Rubio \bgroup et al.\egroup
  }2011]{Gonzalez-RubioETAL:11}
Gonz{\'a}lez-Rubio, Jes{\'u}s, Daniel Ortiz-Mart{\'\i}nez, and Francisco
  Casacuberta.
\newblock 2011.
\newblock An active learning scenario for interactive machine translation.
\newblock In {\em Proceedings of the 13th International Conference on
  Multimodal Interfaces {(ICMI)}}, Barcelona, Spain.

\bibitem[\protect\citename{Gonz{\'a}lez-Rubio \bgroup et al.\egroup
  }2012]{Gonzalez-RubioETAL:12}
Gonz{\'a}lez-Rubio, Jes{\'u}s, Daniel Ortiz-Mart{\'\i}nez, and Francisco
  Casacuberta.
\newblock 2012.
\newblock Active learning for interactive machine translation.
\newblock In {\em Proceedings of the 13th Conference of the European Chapter of
  the Association for Computational Linguistics {(EACL)}}, Avignon, France.

\bibitem[\protect\citename{Green \bgroup et al.\egroup }2014]{GreenETAL:14}
Green, Spence, Sida~I. Wang, Jason Chuang, Jeffrey Heer, Sebastian Schuster,
  and Christopher~D. Manning.
\newblock 2014.
\newblock Human effort and machine learnability in computer aided translation.
\newblock In {\em Proceedings of the 2014 Conference on Empirical Methods in
  Natural Language Processing (EMNLP)}, Doha, Qatar.

\bibitem[\protect\citename{Hassan \bgroup et al.\egroup }2018]{HassanETAL:18}
Hassan, Hany, Anthony Aue, Chang Chen, Vishal Chowdhary, Jonathan Clark,
  Christian Federmann, Xuedong Huang, Marcin Junczys{-}Dowmunt, William Lewis,
  Mu~Li, Shujie Liu, Tie{-}Yan Liu, Renqian Luo, Arul Menezes, Tao Qin, Frank
  Seide, Xu~Tan, Fei Tian, Lijun Wu, Shuangzhi Wu, Yingce Xia, Dongdong Zhang,
  Zhirui Zhang, and Ming Zhou.
\newblock 2018.
\newblock Achieving human parity on automatic chinese to english news
  translation.
\newblock {\em CoRR}, abs/1803.05567.

\bibitem[\protect\citename{Hokamp and Liu}2017]{HokampLiu:17}
Hokamp, Chris and Qun Liu.
\newblock 2017.
\newblock Lexically constrained decoding for sequence generation using grid
  beam search.
\newblock In {\em ACL}, Vancouver, Canada.

\bibitem[\protect\citename{Karimova \bgroup et al.\egroup
  }2018]{KarimovaETAL2018}
Karimova, Sariya, Patrick Simianer, and Stefan Riezler.
\newblock 2018.
\newblock A user-study on online adaptation of neural machine translation to
  human post-edits.
\newblock {\em Machine Translation}, 32(4):309--324.

\bibitem[\protect\citename{Kingma and Ba}2014]{kingma2014adam}
Kingma, Diederik~P and Jimmy Ba.
\newblock 2014.
\newblock Adam: A method for stochastic optimization.
\newblock {\em CoRR}, abs/1412.6980.

\bibitem[\protect\citename{Knowles and Koehn}2016]{KnowlesKoehn:16}
Knowles, Rebecca and Philipp Koehn.
\newblock 2016.
\newblock Neural interactive translation prediction.
\newblock In {\em North American component of the International Association for
  Machine Translation (AMTA)}, Austin, {TX, USA}.

\bibitem[\protect\citename{Koehn \bgroup et al.\egroup }2007]{koehn2007moses}
Koehn, Philipp, Hieu Hoang, Alexandra Birch, Chris Callison-Burch, Marcello
  Federico, Nicola Bertoldi, Brooke Cowan, Wade Shen, Christine Moran, Richard
  Zens, et~al.
\newblock 2007.
\newblock Moses: Open source toolkit for statistical machine translation.
\newblock In {\em Proceedings of the 45th Annual Meeting of the Association for
  Computational Linguistics Companion Volume Proceedings of the Demo and Poster
  Sessions (ACL Demo)}, Prague, Czech Republic.

\bibitem[\protect\citename{Kreutzer \bgroup et al.\egroup
  }2017]{KreutzerETAL:17}
Kreutzer, Julia, Artem Sokolov, and Stefan Riezler.
\newblock 2017.
\newblock Bandit structured prediction for neural sequence-to-sequence
  learning.
\newblock In {\em Proceedings of the 55th Annual Meeting of the Association for
  Computational Linguistics {(ACL)}}, Vancouver, Canada.

\bibitem[\protect\citename{Kreutzer \bgroup et al.\egroup
  }2018]{KreutzerETAL2018}
Kreutzer, Julia, Joshua Uyheng, and Stefan Riezler.
\newblock 2018.
\newblock Reliability and learnability of human bandit feedback for
  sequence-to-sequence reinforcement learning.
\newblock In {\em Proceedings of the 56th Annual Meeting of the Association for
  Computational Linguistics (ACL)}.

\bibitem[\protect\citename{Lam \bgroup et al.\egroup }2018]{LamETAL:18}
Lam, Tsz~Kin, Julia Kreutzer, and Stefan Riezler.
\newblock 2018.
\newblock A reinforcement learning approach to interactive-predictive neural
  machine translation.
\newblock In {\em Proceedings of the 21st Annual Conference of the European
  Association for Machine Translation (EAMT)}, Alicante, Spain.

\bibitem[\protect\citename{Marie and Max}2015]{MarieMax:15}
Marie, Benjamin and Aur{\'e}lien Max.
\newblock 2015.
\newblock Touch-based pre-post-editing of machine translation output.
\newblock In {\em Proceedings of the Conference on Empirical Methods in Natural
  Language Processing {(EMNLP)}}, Lisbon, Portugal.

\bibitem[\protect\citename{Michel and Neubig}2018]{MichelNeubig:18}
Michel, Paul and Graham Neubig.
\newblock 2018.
\newblock Extreme adaptation for personalized neural machine translation.
\newblock In {\em Proceedings of the 56th Annual Meeting of the Association for
  Computational Linguistics (ACL)}, Melbourne, Australia.

\bibitem[\protect\citename{Mnih \bgroup et al.\egroup }2016]{MnihETAL:16}
Mnih, Volodymyr, Adri{\`{a}}~Puigdom{\`{e}}nech Badia, Mehdi Mirza, Alex
  Graves, Timothy~P. Lillicrap, Tim Harley, David Silver, and Koray
  Kavukcuoglu.
\newblock 2016.
\newblock Asynchronous methods for deep reinforcement learning.
\newblock In {\em Proceedings of the 33rd International Conference on Machine
  Learning {(ICML)}}, New York, {NY}.

\bibitem[\protect\citename{Nguyen \bgroup et al.\egroup }2017]{NguyenETAL:17}
Nguyen, Khanh, Hal Daum\'{e}, and Jordan Boyd-Graber.
\newblock 2017.
\newblock Reinforcement learning for bandit neural machine translation with
  simulated feedback.
\newblock In {\em Proceedings of the Conference on Empirical Methods in Natural
  Language Processing {(EMNLP)}}, Copenhagen, Denmark.

\bibitem[\protect\citename{Ortiz-Mart{\'i}nez \bgroup et al.\egroup
  }2010]{Ortiz-MartinezETAL:10}
Ortiz-Mart{\'i}nez, Daniel, Ismael Garc{\'i}a-Varea, and Francisco Casacuberta.
\newblock 2010.
\newblock Online learning for interactive statistical machine translation.
\newblock In {\em Human Language Technologies: The 2010 Annual Conference of
  the North American Chapter of the Association for Computational Linguistics
  {(NAACL-HLT)}}, Los Angeles, {CA}.

\bibitem[\protect\citename{Papineni \bgroup et al.\egroup
  }2002]{papineni2002bleu}
Papineni, Kishore, Salim Roukos, Todd Ward, and Wei-Jing Zhu.
\newblock 2002.
\newblock Bleu: a method for automatic evaluation of machine translation.
\newblock In {\em Proceedings of the 40th Annual Meeting of the Association for
  Computational Linguistics (ACL)}, Philadelphia, {PA, USA}.

\bibitem[\protect\citename{Peris and Casacuberta}2018]{PerisCasacuberta:18}
Peris, {\'{A}}lvaro and Francisco Casacuberta.
\newblock 2018.
\newblock Active learning for interactive neural machine translation of data
  streams.
\newblock In {\em Proceedings of the 22nd Conference on Computational Natural
  Language Learning (CoNLL)}, Brussels, Belgium.

\bibitem[\protect\citename{Peris \bgroup et al.\egroup }2017]{PerisETAL2017}
Peris, {\'{A}}lvaro, Luis Cebri{\'{a}}n, and Francisco Casacuberta.
\newblock 2017.
\newblock Online learning for neural machine translation post-editing.
\newblock {\em CoRR}, abs/1706.03196.

\bibitem[\protect\citename{Popovi{\'c}}2015]{popovic2015chrf}
Popovi{\'c}, Maja.
\newblock 2015.
\newblock chrf: character n-gram f-score for automatic mt evaluation.
\newblock In {\em Proceedings of the Tenth Workshop on Statistical Machine
  Translation}, Lisbon, Portugal.

\bibitem[\protect\citename{Post and Vilar}2018]{PostVilar:18}
Post, Matt and David Vilar.
\newblock 2018.
\newblock Fast lexically constrained decoding with dynamic beam allocation for
  neural machine translation.
\newblock In {\em Proceedings of the 2018 Conference of the North American
  Chapter of the Association for Computational Linguistics: Human Language
  Technologies (NAACL-HLT)}, New Orleans, LA, {USA}.

\bibitem[\protect\citename{Ross \bgroup et al.\egroup }2011]{RossETAL:11}
Ross, St{\'e}phane, Geoffrey~J. Gordon, and J.~Andrew Bagnell.
\newblock 2011.
\newblock A reduction of imitation learning and structured prediction to
  no-regret online learning.
\newblock In {\em Proceedings of the fourteenth international conference on
  artificial intelligence and statistics (AISTATS)}, Fort Lauderdale, {FL,
  USA}.

\bibitem[\protect\citename{Settles and Craven}2008]{SettlesCraven:08}
Settles, Burr and Mark Craven.
\newblock 2008.
\newblock An analysis of active learning strategies for sequence labeling
  tasks.
\newblock In {\em Proceedings of the Conference on Empirical Methods in Natural
  Language Processing {(EMNLP)}}, Honolulu, Hawaii.

\bibitem[\protect\citename{Sutton and Barto}2018]{SuttonBarto:18}
Sutton, Richard~S. and Andrew~G. Barto.
\newblock 2018.
\newblock {\em Reinforcement Learning. An Introduction}.
\newblock The {MIT} Press, second edition.

\bibitem[\protect\citename{Sutton \bgroup et al.\egroup
  }2000]{sutton2000policy}
Sutton, Richard~S, David~A McAllester, Satinder~P Singh, and Yishay Mansour.
\newblock 2000.
\newblock Policy gradient methods for reinforcement learning with function
  approximation.
\newblock In {\em Advances in Neural Information Processings Systems (NIPS)},
  Denver, {CO, USA}.

\bibitem[\protect\citename{Turchi \bgroup et al.\egroup }2017]{TurchiETAL:2017}
Turchi, Marco, Matteo Negri, M.~Amin Farajian, and Marcello Federico.
\newblock 2017.
\newblock Continuous learning from human post-edits for neural machine
  translation.
\newblock {\em The Prague Bulletin of Mathematical Linguistics (PBML)},
  108(1):233--244, jun.

\bibitem[\protect\citename{Williams}1992]{Williams:92}
Williams, Ronald~J.
\newblock 1992.
\newblock Simple statistical gradient-following algorithms for connectionist
  reinforcement learning.
\newblock {\em Machine Learning}, 8:229--256.

\bibitem[\protect\citename{Wu \bgroup et al.\egroup }2016]{WuETAL:16}
Wu, Yonghui, Mike Schuster, Zhifeng Chen, Quoc~V. Le, Mohammad Norouzi,
  Wolfgang Macherey, Maxim Krikun, Yuan Cao, Qin Gao, Klaus Macherey, Jeff
  Klingner, Apurva Shah, Melvin Johnson, Xiaobing Liu, Lukasz Kaiser, Stephan
  Gouws, Yoshikiyo Kato, Taku Kudo, Hideto Kazawa, Keith Stevens, George
  Kurian, Nishant Patil, Wei Wang, Cliff Young, Jason Smith, Jason Riesa, Alex
  Rudnick, Oriol Vinyals, Greg Corrado, Macduff Hughes, and Jeffrey Dean.
\newblock 2016.
\newblock Google's neural machine translation system: Bridging the gap between
  human and machine translation.
\newblock {\em CoRR}, abs/1609.08144.

\bibitem[\protect\citename{Wuebker \bgroup et al.\egroup }2016]{WuebkerETAL:16}
Wuebker, Joern, Spence Green, John DeNero, Sasa Hasan, and Minh-Thang Luong.
\newblock 2016.
\newblock Models and inference for prefix-constrained machine translation.
\newblock In {\em Proceedings of the 54th Annual Meeting of the Association for
  Computational Linguistics (ACL)}, Berlin, Germany.

\bibitem[\protect\citename{Wuebker \bgroup et al.\egroup }2018]{WuebkerETAL:18}
Wuebker, Joern, Patrick Simianer, and John DeNero.
\newblock 2018.
\newblock Compact personalized models for neural machine translation.
\newblock In {\em Proceedings of the 2018 Conference on Empirical Methods in
  Natural Language Processing (EMNLP)}, Brussels, Belgium.

\end{thebibliography}
\bibliographystyle{mtsummit2019}

\end{document}